# CMed-GPT: Prompt Tuning for Entity-Aware Chinese Medical Dialogue Generation

Zhijie Qu[1], Juan Li[1], Zerui Ma[1], Jianqiang Li[1]

[1] Beijing University of Technology, Beijing 100124, China

**Abstract.** Medical dialogue generation relies on natural language generation techniques to enable online medical consultations. Recently, the widespread adoption of large-scale models in the field of natural language processing has facilitated rapid advancements in this technology. Existing medical dialogue models are mostly based on BERT and pre-trained on English corpora, but there is a lack of high-performing models on the task of Chinese medical dialogue generation. To solve the above problem, this paper proposes CMed-GPT, which is the GPT pre-training language model based on Chinese medical domain text. The model is available in two versions, namely, base and large, with corresponding perplexity values of 8.64 and 8.01. Additionally, we incorporate lexical and entity embeddings into the dialogue text in a uniform manner to meet the requirements of downstream dialogue generation tasks. By applying both fine-tuning and p-tuning to CMed-GPT, we lowered the PPL from 8.44 to 7.35. This study not only confirms the exceptional performance of the CMed-GPT model in generating Chinese biomedical text but also highlights the advantages of p-tuning over traditional fine-tuning with prefix prompts. Furthermore, we validate the significance of incorporating external information in medical dialogue generation, which enhances the quality of dialogue generation.

## 1 Introduction

In the context of the current global health crisis, telemedicine's role as a supplement to traditional healthcare has grown in significance. Telemedicine can not only help address the imbalance in the distribution of medical resources but also alleviate the problem of resource scarcity. Additionally, it can enhance medical efficiency and convenience, facilitate real-time communication between patients and doctors, reduce treatment duration, and improve overall medical efficiency. Due to the rapid advancement of Artificial Intelligence (AI) systems, an increasing number of medical researchers are eager to advance intelligent AI dialogue systems into virtual medical practitioners. These virtual medical practitioners can engage in patient consultations, understand their medical conditions and complete medical histories, and offer well-informed clinical recommendations. Consequently, in recent years, He [1], Liu [2], Li [3], Wei [4], Xu [5] and others have proposed task-oriented medical dialogue models to play the role of virtual medical practitioners and engage in one-on-one consultations with patients. However, the presence of specialized phrases and formal medical expressions in Chinese medical conversations makes medical dialogue systems more difficult to implement than task-oriented dialogue systems (TDS).

Pre-training followed by fine-tuning has emerged as the prevailing approach for transferring the capabilities of large models to downstream tasks within the NLP domain [6], including medical dialogue generation. In the biomedical field, researchers have begun investigating the application of pretrained language models. Unfortunately, general pretrained models perform poorly in the medical domain [7, 8], likely due to the presence of a significant number of domain-specific terms that are difficult for these models to comprehend. To enhance pretrained model performance in the medical domain, a series of expert models based on medical datasets have emerged. Even without fine-tuning, PubMedBERT [8], BioBERT [9], SciBERT [10], and DilBERT [11] models outperform BERT in downstream tasks. This represents an outstanding development, as researchers appear to have identified the future research direction for telemedicine: medical pre-training models and fine-tuning.

However, most prior research has focused on BERT-pretrained models and English medical dialogue datasets, leading to underwhelming performance in the generation of Chinese medical dialogues. After analyzing the causes, the main problems with the existing methods are as follows: Regarding medical dialogue datasets, most of the dialogue datasets are in English not Chinese. Regarding models, BERT, a representative bidirectional language model with its bidirectional attention mechanism, may not be as suitable as unidirectional models like GPT for natural language generation (NLG) tasks. Regarding medical domain knowledge, it is more specialized compared to other domains, characterized by complex, less frequent vocabulary, and abundant technical jargon. Undoubtly, these three factors pose significant challenges for the models.

This paper makes the following major contributions to solving the above problems:

1. We collect medical dialogue datasets in Chinese and proceed with pre-training of GPT on the Chinese medical dialogue datasets to develop a Chinese generalized pre-trained language model, namely CMed-GPT, for the biomedical domain.
2. To enhance the model's understanding of medical dialogue, we executed fine-tuning and p-tuning on the pre-trained CMed-GPT model employing downstream medical dialogue data. This enhanced the model's comprehension while simultaneously reducing the number of parameters during the fine-tuning stage.
3. To enable the model to better understand the entities in the medical domain, we combine lexical and entity embeddings with dialog history to obtain the entity-aware medical dialog generation models KB-FT-CMed-GPT and KB-PT-CMed-GPT. The results demonstrate the effectiveness of our method.

The paper is organized as follows: the related works on dialogue generation is shown in Sect. 2. Datasets are shown in Sect. 3. Details of the proposed Chinese medical dialogue generation model is introduced in Sect. 4. Experimental and results are presented in Sect. 5. Finally, conclusion is given in Sect. 6.

## 2 Related Work

For recent research on dialogue generation, pre-training on large-scale datasets and fine-tuning on downstream tasks have proved to be a successful paradigm and have become the standard models. Two of the typical models are BERT and GPT.

BERT is a bidirectional encoder representation language model based on the transformer architecture.In subsequent studies, the improvement points of BERT are mainly focused on two parts: the construction of large-scale corpus data [1] and model pre-training [12]. In the field of biomedicine, scholars have proposed several approaches to enhance the performance of BERT models by pre-training them with English medical domain datasets. These approaches include the following three ideas: 1) Continued pre-training using medical domain data, such as BioBERT [9] and DilBERT [11]; 2) Pre-training from scratch using medical domain data, such as SciBERT [10] and PubMedBERT [8]; 3) Pre-training based on self-supervised tasks specifically designed for the medical domain, such as MC-BERT [13], SMedBERT [14] and BERT-MK [15]. While various biomedical pre-trained language models based on BERT have achieved great success in natural language understanding and classification tasks, few scholars have worked on the task of generating BERT models due to its Transformer encoder model architecture that limits NLG.

GPT is an autoregressive language model based on Transformer's decoder. Compared to the BERT, GPT has significantly larger training corpora and model parameters such as GPT-2 [16] and GPT-3 [17]. The modeling structure of GPT makes it outstanding for generative tasks. With the emergence of ChatGPT, a new fine-tuning paradigm based on pre-trained language models, p-tuning, has become a major mainstream approach to improve the performance of GPT on downstream tasks. For domain-specific generative tasks, the inclusion of prompt allows the model to generate text that is more consistent with the user's intent. In the field of biomedicine, while GPT is well-suited for medical dialogue generation, there are very few pretrained language models based on GPT for medical text generation. Previous work on pretraining GPT in the biomedical literature is DARE [18]. However, they pretrained GPT on an extremely limited dataset consisting of only 0.5 million PubMed abstracts and used it solely for data augmentation in relation extraction tasks. Generative pre-trained language modeling for Chinese is a research gap. With the increase of Chinese medical dialogue datasets in recent years, it is a matter of concern how the GPT model performs in the field of Chinese medical dialogue generation.

Therefore, we propose the CMed-GPT specifically designed for Chinese medical dialogue scenarios. During the pre-training phase of the model, we continue to train the GPT model using Chinese medical datasets. In the fine-tuning phase, we perform p-tuning of CMed-GPT by integrating lexical and entity information into the dialogue text in a uniform manner. This approach not only further demonstrates the effectiveness of domain-specific pretraining on improving model performance but also significantly reduces the parameter count compared to fine-tuning. Furthermore, the uniform incorporation of lexical and entity information in this paper, as opposed to the discrete approach, enhances the model's understanding ability and generates higher-quality dialogue responses that better align with user intent.

## 3 Datasets

Datasets are crucial for pre-trained language models. This section aims to provide a comprehensive description of the datasets utilized in this study. The datasets used for model pre-training include Chinese medical dialogue datasets and medical books.

**Chinese Medical Dialogue Datasets.** The dataset consists of seven single and multi-turn medical dialogue datasets from different sources. It includes five multi-turn medical dialogue datasets labeled with entities such as symptoms, diseases, drugs, examinations, etc., as well as two medical dialogue datasets without entity labeling. The above datasets were obtained from online consulting medical websites, smart conversation clinic competitions and previous studies. The data is shown in Table 1:

**Table 1.** Chinese medical dialogue datasets.

| Dataset | Diseases | Dialogues | Utterances | Tokens | Entities |
|---|---|---|---|---|---|
| CHIP-MDCFNPC | - | 8000 | 247,520 | 4259,819 | 5494,944 |
| CMDD | 4 | 2064 | 86,874 | 868,738 | 651,553 |
| cMQA-master | - | 260,000 | 520,000 | 67,080,000 | - |
| IMCS-IR | 6 | 3052 | 122,080 | 1,599,248 | 1,903,227 |
| MedDG | 12 | 17,864 | 385,862 | 6829,764 | 4692,086 |
| ChunYu | 15 | 12,842 | 317,197 | 3,362,292 | 4091,846 |
| COVID-DDC | 1 | 1088 | 9465 | 405,128 | - |

**Chinese Medical Book Text Dataset.** To enhance the comprehension capabilities of pre-trained models, this paper additionally incorporates open-source medical book text data. The datasets were used to further pre-train existing models, resulting in a pre-trained Chinese language model for the medical field. The book text encompasses pharmacology, diagnostics, pathology, etc., totaling approximately 655.8MB of Chinese biomedical text and roughly 11.2 million tokens.

## 4 Method

### 4.1 Pre-Training Model

**Model Structure.** We primarily continue pretraining on GPT2-Chinese[1], which follows the structure of a Transformer decoder [19] (GPT structure) and employs a multi-head unidirectional attention mechanism. Given the dialogue history $X = X_1, X_2, X_3...X_i...X_k$, where $X_i$ is either a doctor's or a patient's utterance. The utterance $X_i = [u_1^i, u_2^i \ldots u_j^i \ldots u_s^i]$ has s tokens, when predicting the token $u_j^i$, the model will mask the tokens after $u_j^i$. The model structure (see Fig. 1).

---

[1] https://github.com/Morizeyao/GPT2-Chinese

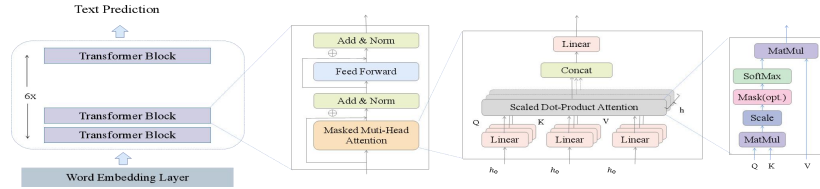

**Fig. 1.** Pre-trained language model structure.

As shown in Fig. 1, the model structure is divided into three layers, including the word embedding layer, transformer block layer, and text prediction layer. The GPT model architecture consists of six transformer block layers, with its core component being the multi-head attention structure. The input matrix $h_0$ is transformed using linear transformation matrices $WQ/WK/WV$ to obtain $Q$, $K$, and $V$, which are used to compute the output of the self-attention mechanism.

$$head_i(Q,K,V) = soft\,max(\frac{Q_iK_i^T}{\sqrt{d_k}})V \qquad (1)$$

Multi-Head Attention consists of multiple self-attention layers:

$$Multihead(Q,K,V) = Concat(head_1, head_2, \ldots, head_k)W \qquad (2)$$

**Model Training Objectives.** Given a dialogue history $X = X_1, X_2, X_3 \ldots X_i \ldots X_k$, the objective of training the model is to minimize the following likelihood function:

$$min\left(-\frac{1}{K}\sum_{i=1}^{K}\sum_{j=1}^{S}log\,P(u_j^i \mid u_i^i \ldots u_{j-1}^i)\right) \qquad (3)$$

The final loss is:

$$Loss = CrossEntorpyLoss(H) \qquad (4)$$

Where $H$ represents the vector after passing through the text prediction layer.

In the following, fine-tuning, p-tuning, and the model with lexical and entity embeddings, the training objectives for these models all involve the next token prediction task, and the formulas are as shown in Eq. 4.

### 4.2 Medical Dialogue Generation Model

**Fine-tuning.** Fine-tuning is an effective method for transferring the capabilities of pretrained models to downstream tasks. Therefore, this paper incorporates fine-tuning as one of the techniques to evaluate the performance of CMed-GPT. Prior studies have extensively employed fine-tuning on models such as BERT or GPT [20,21,22], and this paper refrains from delving further into this topic.

**P-tuning.** Compared to fine-tuning, p-tuning [23, 24] necessitates a smaller amount of data, fewer model parameters, and thus, reduces the demand on GPU memory. This renders it an essential technique for expeditiously transferring pretrained models to downstream tasks.

For p-tuning, we keep the parameters $\varphi$ of the pre-training fixed and add discrete prefix prompt $Prefix = \{p_1, p_2 \dots p_m\}$, before the dialogue text to obtain $Z = [Prefix; X; Y]$. The goal of the model fine-tuning is to maximize the likelihood of $Y$, $P_{\varphi\theta_p} = P(Y|[Prefix; X])$, where the parameter $\theta_p \ll \varphi$. Assuming that the pre-trained model has a vocabulary size of $V$, $V_p$ prompt tokens, and a hidden layer dimension of $H$, the model's word embedding matrix would be of size $E \in V \times H$, and the prompt's word embedding matrix would also be of size $E_p \in V_p \times H$. Considering the input $X = X_1, X_2, X_3 \dots X_i \dots X_k$ comprising utterances $X_i = [u_1^i, u_2^i \dots u_j^i \dots u_s^i]$, where $u_j^i \in V$, and the prompt prefix $Prefix = \{p_1, p_2 \dots p_m\}$, where $p_m \in V_p$. During the training process, the prompt tokens are concatenated with the input text before forwarding it. This concatenation transforms the input $X$ into $Prefix, X_1, X_2, X_3 \dots X_i \dots X_k$. The loss is then computed based on this modified input text. It is important to note that the model's parameter gradients are set to false, while only the prompt parameter gradients are set to true. As a result, during the backward pass, the gradients for the model parameters are set to 0, while only the prompt parameters are updated.

**Lexical and Entity Embeddings.** In the medical domain, there are many specialized and less common entities (long-tail entities) that are relatively rare in general corpora. This rarity can lead to inaccurate word vector representations for these entities. By enhancing the representation of entities, it helps the model better understand the context of the conversation and improves the accuracy of the model's responses [25].In this paper, we represent the entity positions using one-hot vectors. Consequently, the entity information $E_i$ is encoded as an entity vector.

$$E_i = e_1^i, \dots, e_s^i \tag{5}$$

Where s represents the length of a utterance，$e_j^i = 0|1$.

In addition to one-hot encoding entity positions, we also enhance the text representation using lexical embedding. In general, entities in medical text are predominantly nouns (n), while the severity of symptoms is primarily adjectives (adj). These lexical have significant importance in medical text. Therefore, we use "jieba" for lexical tagging in the text, identifying nouns, adjectives, and verbs and assigning them numerical labels, such as nouns: 0, adjectives: 1, verbs: 2. These labels are encoded into vectors and introduced into the text vectors. The modified model structure is shown in Fig. 2. Compared to the original structure, we use a positional encoding-like approach to embed lexical and entity encodings within the input embedding. This modification is intended to enhance the model's ability to comprehend medical text.

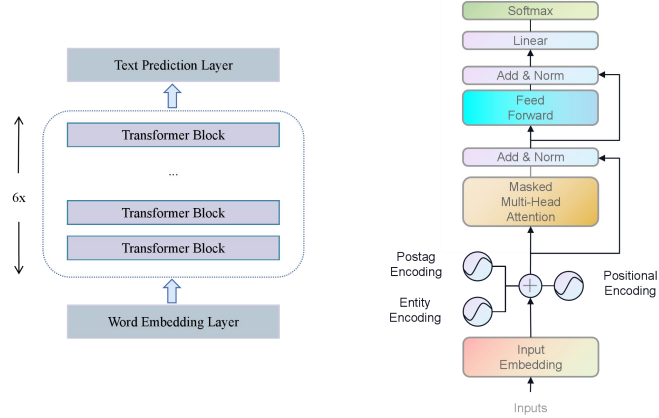

**Fig. 2.** model structure for fusion entity and lexical embeddings.

Assuming the original word vector representation is $E_w$ and the position vector is $E_p$ the input vector for the block is $E = E_w + E_p$, After adding the lexical vector $E_t$ and the entity vector $E_e$:

$$E = E_w + E_p + E_t + E_e \tag{6}$$

## 5    Experiments

In this section, we present the experimental setup and related experimental results for the pre-training model CMed-GPT and fine-tuning, p-tuning.

### 5.1    Experimental Setting

The model utilizes all the data from Section 3 for pre-training, excluding CMDD and IMCS-IR, which is divided into training and test sets at a ratio of 100:1. The AdamW [26] training optimizer was employed with β1 and β2 values set to 0.9 and 0.95, respectively. Weight decay was set at 0.1, and gradient norm clipping was performed at 0.5. A warmup cosine scheduler was used with an initial learning rate of 0.0001, a warmup step of 2000, followed by a cosine decay to a minimum learning rate of 5e-6. The minimum learning rate was maintained after 100,000 steps and continued training until completion, totaling 3 epochs. The maximum text length was set to 512, and if the text exceeded this length, it was truncated from the end to 512. The batch size was set at 32. The entire model was trained using float16 mixed precision, which accelerates training speed while maintaining accuracy compared to float32. Lastly, the base model was trained for 20 days on a 4-card machine with NVIDIA TITAN RTX 24GB, while the large model was trained for 30 days.

To evaluate the model's performance in downstream tasks, we conducted p-tuning/fine-tuning using CMDD and IMCS-IR2 data. The dataset was partitioned into training and test sets, with a ratio of 8:2. The hyperparameters for fine-tuning remained consistent with the pretraining phase, except for a modification in the initial learning rate, set to 5e-5. The model underwent training for 6 epochs. For p-tuning, different quantities of prompt tokens {1, 25, 50, 75, 100} were employed in experimental trials.

## 5.2 Experimental Results

**Pre-training Model.** The pretrained language model for the Chinese medical domain proposed in this paper hasn't found an equivalent pretrained model. Here, we compared it with four other models: Bert-base-chinese[2], Chinese-GPT2-base[3], Chinese-GPT2-large[3], and Medbert[4]. Bert-base-chinese: A BERT model pretrained on general Chinese text data; Chinese-GPT2: A GPT model pretrained on general Chinese text data; Medbert: A BERT model pretrained on Chinese medical domain text data.

Given the nature of the task being a language modeling task, perplexity (PPL) was employed as the metric for evaluating the model's performance. The formula for calculating PPL is as follows:

$$PPL = e^{loss}, e \approx 2.78 \tag{7}$$

The results are shown in the Table 3, and it was observed that general models, whether BERT or GPT2, perform poorly on Chinese medical text, with PPL values greater than 20. However, Medbert, which has been fine-tuned on medical text, outperforms general BERT models, further emphasizing the importance of continued training on medical domain text.

**Table 3.** PPL of CMed-GPT compared to other models in the medical domain test set

| Model | Parameters | Test PPL | Size |
|---|---|---|---|
| Bert-base-chinese[2] | 6 layers，12 head，768 hidden size | 22.63 | 110M |
| Chinese-GPT2-base[3] | 6 layers，12 head，768 hidden size | 20.82 | 110M |
| Chinese-GPT2-large[3] | 12 layers，12 head，768 hidden size | 18.64 | 200M |
| Medbert[4] | 6 layers，12 head，768 hidden size | 9.37 | 110M |
| CMed-GPT-base(Ours) | 6 layers，12 head，768 hidden size | 8.64 | 110M |
| CMed-GPT-Large(Ours) | 12 layers，12 head，768 hidden size | 8.01 | 200M |

In addition, CMed-GPT-base also outperforms an equivalently sized Medbert. This can be attributed to the inherent suitability of unidirectional language models for generation tasks. Furthermore, increasing the model size (from 6 to 12 layers) leads to a noticeable improvement in performance.

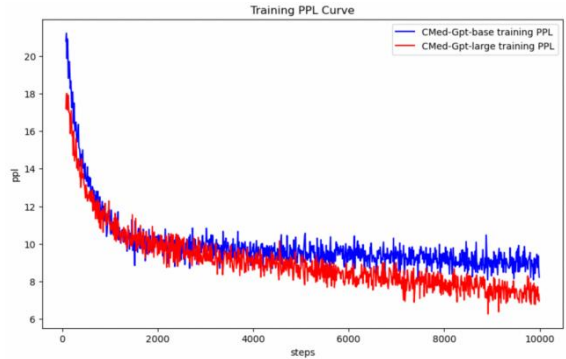

**Fig. 3.** Train PPL curves for CMed-Gpt-base/large

**Fine-tuning/P-tuning.** In the fine-tuning and p-tuning phases, we utilized CMDD dataset and MCS-IR dataset. The results are shown in Table 4, where CMed-GPT-base is the pre-trained model, KB-FT-CMed-GPT is the CMed-GPT-base post-fine-tuning model and KB-PT-CMed-GPT is the CMed-GPT-base post-p-tuning model. It was observed that after fine-tuning/p-tuning on the downstream dialog task datasets, the PPL decreased from above 8 to below 8, indicating an improvement of approximately 10%. Additionally, p-tuning achieves comparable or slightly superior results compared to fine-tuning. This suggests that the prefix prompts used in p-tuning had a significant impact. Compared to fine-tuning, p-tuning required fewer parameters and less time for training.

**Table 4.** CMed-GPT-base fine-tuning/p-tuning results on CMDD and IMCS-IR.

| Dataset | CMed-GPT-base | KB-FT-CMed-GPT | KB-PT-CMed-GPT |
|---------|---------------|----------------|----------------|
| CMDD    | 8.44          | 7.68           | 7.58           |
| IMCS-IR | 8.79          | 7.75           | 7.62           |

In the p-tuning phase, we experimented with different numbers of prompt tokens to assess their impact on the results. For this experiment, we only used the CMDD dataset, and the results are as follows:

**Table 5.** PPL of the model with different prompt token.

| Dataset | 1    | 25   | 50   | 75   | 100  |
|---------|------|------|------|------|------|
| CMDD    | 8.32 | 7.85 | 7.68 | 7.69 | 7.93 |

Based on the Table 5, it is evident that the best result is achieved at v=50, suggesting that the addition of prompts does not always lead to performance improvements for the model. This phenomenon is likely influenced by the volume of tuning data. When there is a large amount of data for tuning, more parameters are needed to fine-tune and fit the data. However, when the tuning data is limited, an appropriate number of prompt tokens can improve model performance.

**Entity-Aware Medical Dialogue Generation.** The introduction of entity and lexical enhanced GPT models has been widely discussed in other papers as well. However, the approach of incorporating entities often involves identifying them and discretely appending them to the input text[5]. This results in discrete entity vectors that cannot uniformly represent each token. In contrast to other papers, we integrate entity and lexical vectors with the text embedding and positional embedding in an encoded manner, similar to the transformer input format. The uniformly spliced matrix is fed into our model after p-tuning. Experimental results indicate that incorporating entity and lexical information further improves the model's performance.

**Table 6.** PPL of the model after lexical/entity embedding for different datasets.

| Dataset | P-tuning | Lexical | Entity | Lexical and entity | Entity splicing method[5] |
| --- | --- | --- | --- | --- | --- |
| CMDD | 7.58 | 7.5 | 7.46 | 7.35 | 7.89 |
| IMCS-IR | 7.62 | 7.58 | 7.53 | 7.43 | 7.93 |

The inclusion of lexical and entity vectors in medical text has strengthened the representation of certain important words in the text. This makes it easier for the model to generate accurate responses during subsequent conversations.

# 6 Conclusion

In this paper, we first conducted a comprehensive compilation and analysis of existing Chinese medical dialogue datasets and medical book data. Subsequently, using this valuable collection of Chinese medical text data, we successfully pretrained a powerful Chinese medical pre-trained language model, named CMed-GPT. This model not only possesses extensive medical knowledge but also has the ability to understand and generate medical dialogue text.

To address the common long-tail entity problem in medical text, we propose an innovative approach that effectively integrates lexical and entity information during model training. Additionally, we devised discrete pseudo-token prefixes, which were appended to the text data. Through a series of rigorous experiments, we have concluded that our proposed CMed-GPT, including both the base and large versions, exhibits outstanding performance in the medical domain.

---

[5]  https://arxiv.org/abs/2212.06049

Furthermore, our CMed-GPT model can be easily applied to various downstream tasks in the medical field. By performing simple p-tuning, which involves adjusting a small number of parameters, the model's performance on specific tasks can be significantly improved. This provides a powerful and efficient tool for medical information processing.

To summarize, this study not only successfully constructs a pre-trained language model CMed-GPT for the Chinese medical domain, but also proposes an effective method for entity information incorporation and demonstrates its excellent performance in medical text processing tasks. In the future, we will continue to improve the model and explore more application areas to better meet the needs of medical information processing and contribute to advancements in the medical field.